\documentclass[letterpaper, 10 pt, conference]{ieeeconf}  

\IEEEoverridecommandlockouts                              

\title{\LARGE \bf
OpenGS-Fusion: Open-Vocabulary Dense Mapping with Hybrid 3D Gaussian Splatting for Refined Object-Level Understanding
}

\author{ Dianyi Yang$^{1}$, Xihan Wang$^{1}$, Yu Gao$^{1}$, Shiyang Liu$^{1}$, Bohan Ren$^{1}$, Yufeng Yue$^{1}$, Yi Yang$^{*, 1}$
\thanks{This work was partly supported by National Natural Science Foundation of China (Grant No. NSFC 62233002) and National Key R\&D Program of China (2022YFC2603600). (*Corresponding Author: Y. Yang, yang\_yi@bit.edu.cn)}
\thanks{$^{1}$School of Automation, Beijing Institute of Technology, Beijing, China}%
}

\usepackage{amsmath}
\usepackage{color}
\usepackage{amssymb}
\usepackage{graphicx}

\usepackage{float}
\usepackage{booktabs}
\usepackage{stfloats}
\usepackage{multirow}
\usepackage{makecell}
\usepackage{cite}
\usepackage{threeparttable}
\usepackage{colortbl}
\usepackage{xcolor}
\usepackage{xcolor}
\usepackage{graphicx}
\usepackage{multirow}
\usepackage{boldline}
\usepackage{pifont}
\usepackage{url}
\usepackage[symbol]{footmisc}
\usepackage{amsmath}
\usepackage{tcolorbox}
\usepackage{amssymb}
\usepackage{tabularx}

\newcommand{\cmark}{\ding{51}}
\newcommand{\xmark}{\ding{55}}

\usepackage[colorlinks,
            linkcolor=red,
            anchorcolor=green,
            citecolor=green
            ]{hyperref}

\begin{document}

\maketitle
\thispagestyle{empty}
\pagestyle{empty}

\begin{abstract}

Recent advancements in 3D scene understanding have made significant strides in enabling interaction with scenes using open-vocabulary queries, particularly for VR/AR and robotic applications. Nevertheless, existing methods are hindered by rigid offline pipelines and the inability to provide precise 3D object-level understanding given open-ended queries. In this paper, we present OpenGS-Fusion, an innovative open-vocabulary dense mapping framework that improves semantic modeling and refines object-level understanding. OpenGS-Fusion combines 3D Gaussian representation with a Truncated Signed Distance Field to facilitate lossless fusion of semantic features on-the-fly. Furthermore, we introduce a novel multimodal language-guided approach named MLLM-Assisted Adaptive Thresholding, which refines the segmentation of 3D objects by adaptively adjusting similarity thresholds, achieving an improvement 17\% in 3D mIoU compared to the fixed threshold strategy. Extensive experiments demonstrate that our method outperforms existing methods in 3D object understanding and scene reconstruction quality, as well as showcasing its effectiveness in language-guided scene interaction. The code is available at \href{https://young-bit.github.io/opengs-fusion.github.io/}{https://young-bit.github.io/opengs-fusion.github.io/}.

\end{abstract}

\section{INTRODUCTION}
Open-vocabulary scene understanding is critical for enabling various interaction tasks in real-world environments. Recent studies \cite{conceptfusion, lerf2023, open-fusion, opengaussian} have shown that by projecting 2D semantic information extracted from vision language models (e.g., CLIP\cite{openclip}) into 3D scene representations, users can interact with scenes through open-vocabulary queries. This capability is particularly beneficial for VR/AR \cite{gaussian_grouping,VR-GS}, as well as for enhancing complex robotic tasks, including navigation \cite{VLM, clipfields} and object manipulation\cite{lerftogo, gaussiangrasper}.

\begin{figure}
    \centering
    \includegraphics[width=1\linewidth]{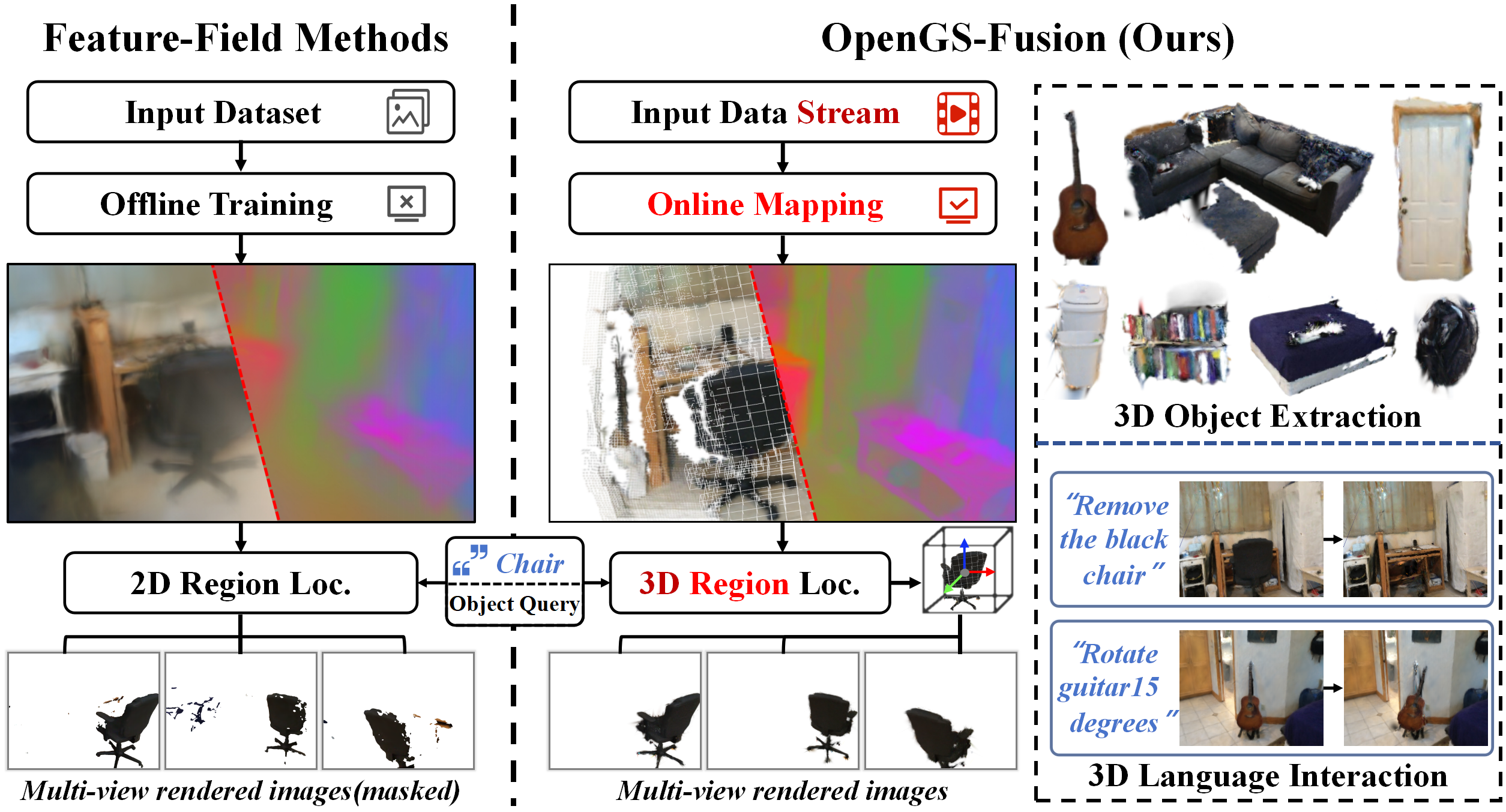}
    \caption{Comparison of Model Architectures. Compared to 3DGS-feature-field-based methods \cite{langsplat,legaussian,feature-3dgs,FMGS}, our approach enables online modeling of scene appearance, geometry, and semantics while supporting 3D object-level queries. Our method enables versatile task-oriented interactions, such as 3D object extraction and editing in an interactive manner. The results on the left are from \cite{langsplat}.}
    \label{Fig. 1}
\end{figure}

A key factor in facilitating these tasks is the underlying scene representation that bridges the gap between 2D and 3D. It should enable timely updates with incoming data while maintaining accurate geometric, appearance, and semantic modeling. Traditional methods like point clouds\cite{conceptfusion, clipfields} and voxel-based approaches\cite{open-fusion, uni-fusion} integrate well with many SLAM systems. However, they lack the ability to synthesize novel views and struggle with high-fidelity reconstruction.\cite{splatam} Although NeRF-based methods \cite{3d-ovs,lerf2023,nef2} mitigated these issues, they remain limited by fixed resolution and inefficient per-pixel raycasting. \cite{LEGS}

Compared to NeRF, Gaussian splatting\cite{3dgs} exhibits significant advantages in training efficiency and explicit scene representation, making it highly suitable for scene understanding tasks. Therefore, numerous studies\cite{langsplat,legaussian,feature-3dgs,tiger,clipgs,FMGS} have explored incorporating semantic features into Gaussian scene representation to construct a language feature field.
Although these methods demonstrate impressive performance in multi-view semantic understanding, several limitations remain:

1) \textbf{Rigid Offline Pipeline.} These methods rely on essential preprocessing steps, such as pretraining 3D Gaussian representation or offline compression of high-dimensional semantic features. However, in real-world applications, for example, robotic exploration and embodied interaction, models must support online perception. The inflexible pipeline of these methods limits their scalability to these applications, as any scene updates or new data necessitate retraining.

 2) \textbf{Limited 3D Object-Level Understanding.} Most existing methods respond to textual queries by highlighting target regions in 2D renderings from arbitrary viewpoints, lacking the ability to provide precise 3D object models. Additionally, these methods use a fixed threshold to filter irrelevant regions during text query processing, which often fails to accurately delineate object boundaries due to the contextual dependency and ambiguity of semantic features.

To address these challenges, we propose OpenGS-Fusion, a dense mapping framework for open-vocabulary object-level scene understanding. \textbf{1) Online Scene Understanding:} OpenGS-Fusion extends cutting-edge GS-based mapping system\cite{gsicp} by combining 3D Gaussian representations with a Truncated Signed Distance Field (TSDF)\cite{vdbfusion}. The TSDF’s voxel-based structure facilitates lossless fusion of semantic features and guides 3D Gaussian initialization, ensuring full semantic preservation while significantly improving scene update efficiency. \textbf{2) Refined 3D Object Understanding:} To achieve precise 3D object localization, we compute semantic similarity across all spatial regions within the semantic space. We further enhance this process with MLLM-Assisted Adaptive Thresholding (AT-MLLM), a mechanism guided by multimodal large language models \cite{gpt}. By adaptively adjusting thresholds based on semantic similarity, AT-MLLM significantly improves object-level understanding accuracy, achieving a 17\% improvement in 3D mIoU.

In summary, our contributions are as follows.

\begin{itemize}
    \item We introduce OpenGS-Fusion, an innovative open-vocabulary dense mapping framework that leverages a hybrid scene representation to concurrently construct the appearance, geometry, and semantic features of a scene in an online manner.
\end{itemize}

\begin{itemize}
    \item  We introduce a novel open-vocabulary query strategy that leverages MLLM to adaptively adjust the similarity threshold, thereby refining the final object localization.
\end{itemize}

\begin{itemize}
    \item  Extensive evaluations on various datasets demonstrate our competitive performance in scene understanding and reconstruction quality, as well as the effectiveness of language-guided scene interaction.
\end{itemize}

\section{RELATED WORKS}

\subsection{Scene Dense Mapping System}

Early dense mapping research focused on accurately representing environments at both metric and topological levels, emphasizing scene geometry and spatial relationships. Recently, with advancements in radiation field techniques \cite{3dgs,nerf} and SLAM algorithms, scene representation has shifted from pure geometry to an integration of appearance, semantics, and localization. Gaussian splatting-based methods (e.g., LoopSplat\cite{loopsplta}, SplaTAM\cite{splatam}, MonoGS\cite{monogs}) have shown strong potential in geometry and appearance modeling, with some enabling real-time reconstruction (e.g., CG-SLAM\cite{cgslam}, GS-Fusion\cite{wei2024gsfusion}, GSICP-SLAM\cite{gsicp}). Other works (e.g., SGS-SLAM \cite{sgsslam}, SemGaussian-SLAM \cite{semgauss}) incorporate semantics for high-quality mapping but rely on annotated datasets and fixed semantic categories, limiting real-world applicability. In this paper, we leverage existing 2D foundation models without additional training, enabling dense semantic mapping in open-world scenarios.

\subsection{Open-Vocabulary Scene Understanding Methods}
Recent advancements in Visual Language Foundation Models (VLFMs), such as CLIP\cite{openclip}, RegionClip\cite{regionclip}, and MaskClip\cite{zhou2022maskclip}, have driven efforts to lift 2D semantic knowledge into 3D for richer scene understanding and interaction.
Methods like OpenScene\cite{Peng2023OpenScene}, Clip-Fields\cite{clipfields}, Open3DIS\cite{nguyen2023open3dis}, and ConceptFusion\cite{conceptfusion} integrate geometric and semantic information into point clouds, enabling 3D scene perception but at a high storage cost. Open-Fusion\cite{open-fusion} and Uni-Fusion\cite{uni-fusion} address this by incorporating voxelized CLIP features and supporting online scene understanding, but their geometric and appearance accuracy is highly sensitive to voxel resolution. 
3D-OVS\cite{3d-ovs} is the first to use NeRF for semantic modeling in open scenes, while LERF\cite{lerf2023} and OV-NeRF\cite{OV-NeRF} improve the accuracy and consistency of semantic field learning. Their subsequent work, LEGS\cite{LEGS} and OV-Mapping\cite{ovmapping}, extend these methods to online scenes through frame-by-frame fusion. However, all of these methods store semantic information in NeRF, which is prone to catastrophic forgetting in large-scale scenes. \cite{LEGS}

Concurrently, methods based on Gaussian feature fields including Feature-3DGS\cite{feature-3dgs}, LEGaussian\cite{legaussian}, FMGS\cite{FMGS}, FastLGS\cite{fastlgs}, TiGER\cite{tiger}, CLIP-GS\cite{clipgs}, and LangSplat\cite{langsplat}, compress 2D CLIP features onto 3D Gaussian primitives, facilitating high-precision pixel-level scene understanding from any viewpoint via multi-view learning. Nevertheless, they lack semantic localization capabilities for 3D objects. OpenGaussian\cite{opengaussian} and EgoFilter\cite{gu2024egolifter} attempt to address this through 3D instance segmentation. However, such a strategy is inefficient and heavily dependent on 3D segmentation accuracy, limiting their scene understanding capacity. Furthermore, all of these methods are limited to offline learning, making them unsuitable for applications that require online updates, such as robot navigation or interactive tasks. In this paper, we combine the hybrid scene representation TSDF-3DGS to simultaneously model the appearance, geometry, and semantic features of the scene in an online, segmentation-free manner. Additionally, our method supports precise 3D object-level localization given natural language queries.

\begin{figure*}
\vspace{0.2em}
    \centering
    \includegraphics[width=0.97\linewidth]{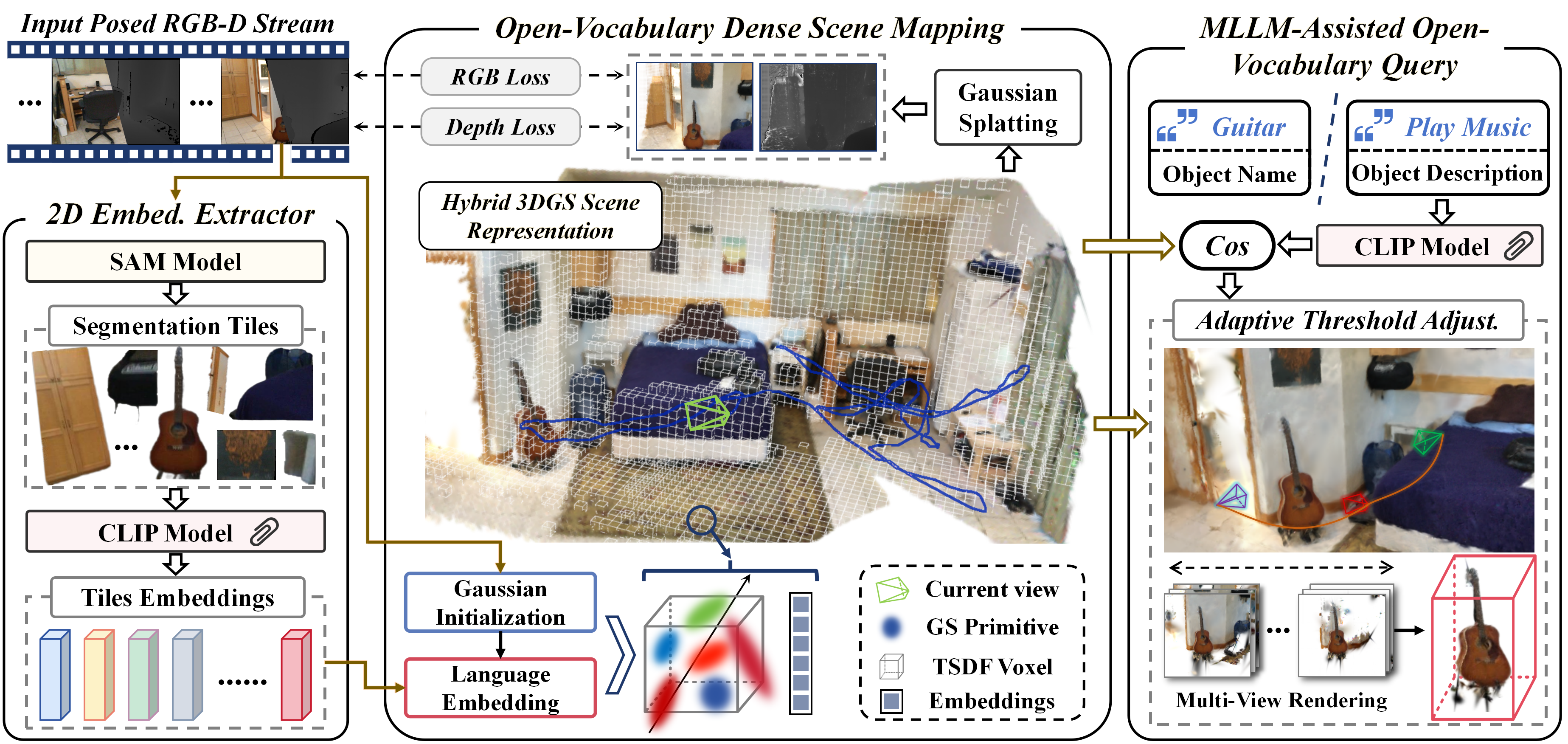}
    \caption{\textbf{Overview of OpenGS-Fusion.} Receiving RGB-D input with 2D language embeddings extracted from 2D foundation models, we simultaneously update the appearance, geometry and semantic features of our hybrid 3D Gaussian scene representation. Additionally, the proposed open-vocabulary query strategy enables precise localization of 3D objects without the need for explicit scene segmentation.}
    \label{overview}
    \vspace{-1.0em}
\end{figure*}

\section{OpenGS-Fusion}

  An overview of our system is shown in Fig.\ref{overview}. Receiving continuous $T$ RGB-D frames of color images $\mathbf{C}_t \in \mathbb{R}^{H \times W \times 3}$ and depth images $\mathbf{D}_t \in \mathbb{R}^{H \times W}$ with poses $\mathbf{P}_t \in \mathbb{R}^{4 \times 4}$.  We first extract 2D semantic features $\mathbf{S}_t$ following the methodology described in Sec.\ref{pixel}. Subsequently, the tuple $\{\mathbf{C}_t, \mathbf{D}_t, \mathbf{S}_t, \mathbf{P}_t\}$ is used to update our hybrid scene representation $\mathcal{M}$ (Sec.\ref{hybrid}), following the approach in Sec.\ref{Open}. Throughout the mapping process, users can leverage our AT-MLLM module (Sec.\ref{Adata}) to perform open-vocabulary queries, enabling the localization of 3D objects of interest and high-quality multi-view rendering.

\subsection{2D Language Embeddings Extractor} \label{pixel}
We adopt a methodology similar to \cite{conceptfusion,langsplat,opengaussian} for extracting language embeddings. Given the RGB image \(\mathbf{C}_t\) from the current view at time \(t\), we first employ a SAM-like model \cite{zhang2023mobilesamv2, sam} to extract a set of region proposals \(\mathcal{R}\). By inputting these proposals into the CLIP model, we can extract region-level semantic features, where all pixels within region $R_i$ share a unified language embeddings $e_i$ and confidence $c_i$. During pixel-level feature querying, the semantic embeddings of each pixel is directly assigned as the embeddings of its corresponding region. Compared to methods\cite{ovmapping,LEGS} that directly output 2D semantic features without using 2D segmentation models, our approach improves the consistency of CLIP features across multiple views by using SAM.

\subsection{Hybrid 3DGS Scene Representation} \label{hybrid}
In general, we represent unknown scenes using an OpenVDB-based TSDF and a 3DGS scene representation. OpenVDB\cite{vdbfusion} is a data structure designed to manipulate unbounded volumetric data, enabling efficient CPU-based operations. It supports flexible data types, such as TSDF and occupancy grids. We adopt TSDF for its ability to perform global geometric perception and integrate 3D semantic embeddings in an efficient manner. Specifically, we configure it as a single resolution grid, where each voxel \( v \) stores a truncated signed distance value \( \phi(v) \) to the nearest surface, with \( \phi(v) \leq 0 \) indicating the voxels located behind the surface. However, TSDF alone cannot capture fine local details or support high-fidelity novel view synthesis. 

Therefore, 3DGS is also chosen as the scene representation because it is differentiable and can be easily projected into 3D space to enable novel view synthesis. A Gaussian primitive can be seen as an ellipsoid, parameterized by 3D center position \( x \), covariance matrix \( \Sigma \), opacity \( o \) and color \( h \). During rendering, Gaussian primitives are sorted by depth and projected from 3D space onto the 2D image plane using the camera pose \( \mathbf{P}_t = \{\mathbf{R}, \mathbf{t}\} \). The covariance matrix \( \Sigma \) is transformed into the 2D plane as:
\begin{equation}
\Sigma' = \mathbf{J} \mathbf{T}^{-1} \Sigma \mathbf{T}^{-T} \mathbf{J}^T,
\end{equation}
where \( \mathbf{J} \) represents the Jacobian of the projection function. The color \( \mathbf{\hat{C}}(p) \) and depth \( \mathbf{\hat{D}}(p) \) of each pixel \( p \) are then computed using alpha-blending, as defined by:

\begin{equation}
\mathbf{\hat{C}}(p) = \sum_{i \in N} h_i \alpha_i \prod_{j=1}^{i-1} (1 - \alpha_j),  \mathbf{\hat{D}}(p) = \sum_{i \in N} d_i \alpha_i \prod_{j=1}^{i-1} (1 - \alpha_j)
\end{equation}
where \( h_i \) and \( d_i \) represent the color and depth of the \( i \)-th Gaussian primitive, respectively. 

In summary, our hybrid scene representation is defined as a set of sparse voxels \( \mathcal{V} = \{v_j\}_{j=1}^{Nv} \), where each voxel \( v_j \) contains the tuple \( \{\omega_j, \phi_j, c_j, \mathcal{G}_j, f_j\} \). Here, \( \omega_j \) denotes the weight, \( \phi_j \) is the TSDF value, \( c_j \) represents the semantic confidence, \( \mathcal{G}_j = \{g_k\}_{k=1}^M \) is the set of Gaussian primitives and \( f_j \) is the semantic embedding within the voxel.
 This hybrid representation leverages the complementary strengths of TSDF and 3DGS. The TSDF provides global structural guidance for initializing Gaussian primitives and stores implicit semantic features, while 3DGS enhances the representation of fine local details and enables high-fidelity multi-view rendering. Further details are discussed in the following section.

\subsection{Open-Vocabulary Dense Scene Mapping} \label{Open}

Given input $\{\mathbf{C}_t, \mathbf{D}_t, \mathbf{S}_t, \mathbf{P}_t\}$, we update the global map $\mathcal{M}$ in three stages. First, it performs TSDF fusion to update the necessary voxels and capture the global scene geometry. We then densify $\mathcal{M}$ with Gaussian primitives within highly occupied voxels to enhance scene representation while updating semantic embeddings. Finally, we incorporate our scene optimization strategy to refine the overall mapping quality in real time. Notably, all input frames participate in the TSDF Fusion process. However, to improve efficiency, the densification and semantic updates processes are executed only when the current frame is identified as a keyframe. We adopt the same keyframe selection strategy as \cite{gsicp}, which will not be detailed here.

\textit{1) VDB-Based TSDF Fusion:} 
We adopt the VDBFusion\cite{vdbfusion} approach to perform TSDF fusion and update the current volumetric map represented by a VDB volume. We first downsample the input $D_t$ and transform the sampled points into a set of rays $\mathcal{R}=\{r_1,...r_N\}$ based on the current pose $\mathbf{P}_t$. Subsequently, we feed these rays into the Differential Digital Analyzer (DDA) provided by the OpenVDB library to obtain all voxels that need updating. Following the original implementation\cite{vdbfusion}, we then calculate the new TSDF value for each voxel $v_k$ and update the final TSDF value $\phi_k$ and the new weight $\omega_k$ accordingly. Mathematically:
\begin{equation}
\begin{aligned}
\omega_k &= \min(\omega_{\text{max}}, \omega_{k-1} + 1), \\
\phi_k &= \frac{\phi_{k-1} \omega_{k-1} + \phi \omega_k}{\omega_{k-1} + \omega_k}
\end{aligned}
\label{blending}
\end{equation}
where $\omega_{\text{max}}$ is an upper limit of weights.

\textit{2) Gaussian Densification and Semantic Fusion: } 
For each keyframe, we initialize our GS primitives using the downsampled points from the previous procedure. Unlike methods that use fixed parameters for initialization \cite{splatam, wei2024gsfusion, monogs}, we adopt a G-ICP based initialization method following \cite{gsicp}. Each point $ x_i $ is transformed into a Gaussian distribution $\{x_i, \Sigma_i\}$ during the G-ICP process. This distribution is then merged with the point's color $ h_i $ and opacity $ o = 0.5 $ to form an initial GS primitive added to the scene. This approach allows our method to obtain a relatively accurate geometric representation at the initialization stage, reducing the optimization cost. Furthermore, in order to enhance robustness against noise and reduce unnecessary GS primitives, we check whether the TSDF value $\phi_k$ of the voxel where each GS resides exceeds a certain threshold \( \gamma \) and whether there are no overlaps with other GS primitives. Only when both conditions are met will it be added to the global map. 
Notably, the voxel $v_j$ of each Gaussian can be queried through the previously mentioned DDA.

Furthermore, for each initialized Gaussian $ g_i $, since it is projected from the $\mathbf{C}_t$, we can obtain its pixel-level embeddings $ e_i $ and confidence $ c_i $. Subsequently, we perform semantic feature fusion based on TSDF as follows:
\begin{equation}
\begin{aligned}
\bar{f}_{j} &= \frac{\bar{c}_j\bar{f}_j + c_i f_i}{\bar{c}_j + c_i}, \\
\bar{c}_j &= \bar{c}_j + c_i
\end{aligned}
\end{equation}
where $\bar{f}_{j}$ and $\bar{c}_j$ represent the semantic feature and confidence of the voxel $v_j$ in which $g_i$ resides.

\textit{3) Scene Optimization Strategy:} To supervise the learning of our Gaussian representation, we apply the same loss function as described in \cite{gsicp}. Furthermore, after introducing each keyframe, we evaluate the TSDF value of every voxel and prune all GS primitives within voxels where the TSDF value is less than $\theta$. This further enhances the efficiency of scene optimization and robustness against noise.

\subsection{ MLLM-Assisted Adaptive Thresholding} \label{Adata}
At any time $t$, given a natural language query $\mathbf{Q}$ containing object names, our objective is to extract the corresponding Gaussian primitives from $\mathcal{M}$. Unlike methods\cite{langsplat,gu2024egolifter,legaussian,FMGS} that focus on locating target objects in 2D rendered images, our approach inherently ensures multi-view consistency of the final results by directly operating in the 3D space.

\begin{figure}
\vspace{0.3em}
    \centering
    \includegraphics[width=0.98\linewidth]{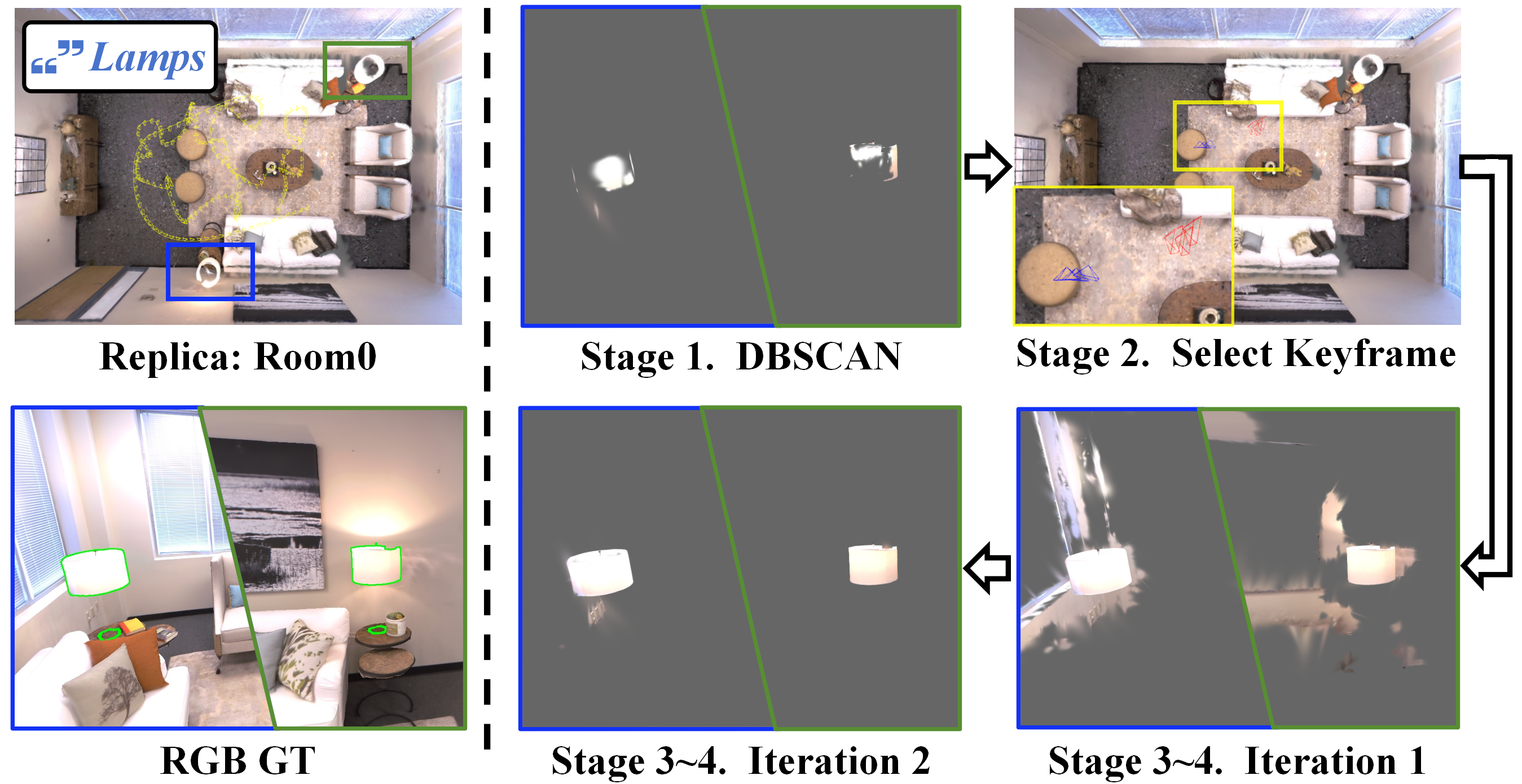}
    \caption{A visual demonstration of our proposed AT-MLLM. This strategy enables the precise localization of multiple objects in 3D space through multi-stage threshold adjustment assisted by MLLM.}
    \label{stages}
     \vspace{-1.5em}
\end{figure}

We first input $\mathbf{Q}$ into the CLIP model to extract text features, which are then compared with semantic features $\mathbf{F}$ of all global voxels $\mathcal{V}$ using cosine similarity. This GPU-accelerated process typically completes in milliseconds. After obtaining the similarities $Sim$, a threshold \( \delta \) is applied to filter the relevant voxels, and all GS points within these voxels.  However, existing methods \cite{langsplat,conceptfusion,FMGS,LEGS,feature-3dgs,legaussian} typically rely on a fixed threshold for filtering, regardless of whether the data is in 2D or 3D. This approach often results in ambiguous object boundaries and challenges in accurately locating spatial objects.

Therefore, we propose an adaptive threshold adjustment strategy assisted by MLLM, where MLLM refers to large vision language models that support both image and text inputs. As shown in Fig.\ref{stages}, the strategy includes four stages. 

\textit{Stage1: } Assuming all the $Sim$ values have been mapped to the range $[0,1]$, we initially set a high threshold $\delta_1$ for filtering, aiming to select the Gaussians set $\mathcal{G}_{1}$ that are highly related to the query. Subsequently, we treat $\mathcal{G}_{1}$ as a point cloud and apply DBSCAN\cite{dbscn} to remove noise points and obtain multiple clusters $\mathcal{B}$, considering that there may be multiple objects matching $\mathcal{Q}$ in the scene.

\textit{Stage2: } For each cluster $\mathcal{B}_l$, we search for $u$ queried frames $\mathcal{W}_l$ from the keyframes based on the shortest distance and the highest visual coverage. The evaluation score for keyframe $w$ is defined as: 
\begin{equation}
\begin{aligned}
E(\mathcal{B}_l, w) = \frac{\text{Cov}(w)}{d(\mathcal{B}_l, w) + \epsilon}
\end{aligned}
\end{equation}
where $d(\cdot)$ represents the spatial distance between the cluster center and the keyframe, and $\text{Cov}(\cdot)$ denotes the visual coverage of the cluster in $w$.

\textit{Stage3: } We initialize a threshold window $TW = [0.5, 1]$ and uniformly sample five thresholds within this range. For each threshold $\delta_q$, we use it to filter the scene primitives, and render a corresponding set of images under $\mathcal{W}_l$ for cluster ${\mathcal{B}_l}$ using Gaussian Splatting.

\textit{Stage4: } These images are fed into the MLLM for evaluation, along with a prompt that contains the user's query $\mathbf{Q}$. The MLLM evaluates the results rendered from each queried frame at the above thresholds to find the best threshold for each $w$ in $\mathcal{W}_l$. By aggregating the results of all frames, we determine the optimal threshold ${\hat{\delta}}$. Finally, we adjust the window $TW = [\hat{\delta} - \xi, \hat{\delta} + \xi]$ and return to Stage3.

After 2-3 iterations, we can obtain the final threshold and the corresponding GS primitives. We demonstrate in Sec.\ref{ablation} that this strategy significantly improves object segmentation accuracy compared to the fixed-threshold strategy.

\section{EXPERIMENT} \label{exp}

\begin{figure}[t]
    \centering
    \vspace{0.5em}
    \includegraphics[width=0.99\linewidth]{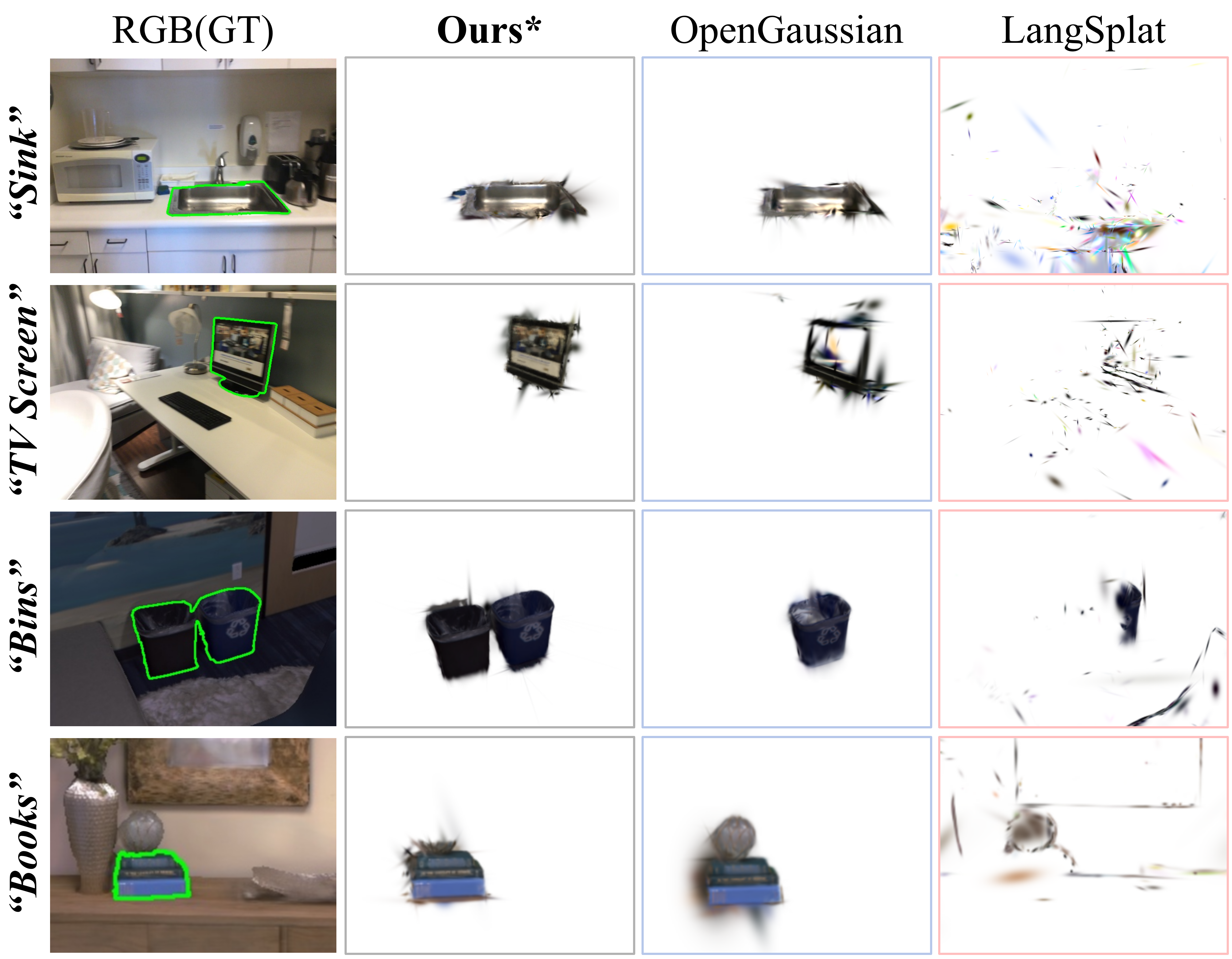}
    \caption{Qualitative comparison of open-vocabulary 3D understanding on ScanNet (top two rows) and Replica (last two rows), benchmarking 3DGS-based methods (OpenGaussian\cite{opengaussian} and LangSplat\cite{langsplat}).}
    \label{exp1}
    \vspace{-0.5em}
\end{figure}

Given an RGB-D video stream, our method enables online scene semantic understanding and high-fidelity map reconstruction. To comprehensively evaluate its performance, we conducted extensive experiments focusing on 3D object-level scene comprehension (Sec. \ref{3D_open}) and dense reconstruction (Sec. \ref{Dense}) capabilities. Furthermore, we demonstrate how the model assists users in language-guided scene editing (Sec. \ref{exp_edit}). Finally, we implement and validate the proposed algorithm in real-world scenarios using our robotic mobile platform (Sec. \ref{real}).

\subsection{3D Open-Vocabulary Scene Understanding.} \label{3D_open}
\noindent\textbf{Settings.} 1) Task: Given an open-ended object-level textual query, we compute the cosine similarity between the query features and the 3D primitive features. Based on similarity scores, we select the most relevant 3D primitives and quantitatively compare them with the ground-truth 3D object segmentation

2) Baselines: We compare our method with 3DGS-based approaches, including LangSplat\cite{langsplat} and OpenGaussian\cite{opengaussian}, as well as points-based scene understanding methods ConceptFusion\cite{conceptfusion} and ConceptGraphs\cite{conceptgraph}. Following \cite{opengaussian}, we extend LangSplat by reconstructing high-dimensional CLIP features from low-dimensional semantic features to perform 3D scene understanding. To ensure fairness, we adhere to \cite{opengaussian} for training LangSplat, OpenGaussian, and our method, without optimizing the positional attributes of Gaussian primitives. 

\vspace{-0.5em}

3) Datasets: The experiments were carried out on 8 Replica \cite{replica19arxiv} scenes and 10 real-world scenes from ScanNetV2 \cite{dai2017scannet}, each of which provides RGB-D images, pose information, and 3D segmentation annotations. Point initialization for the offline methods is derived from the original datasets. 

4) Evaluation Metrics: We consider all foreground objects in each scene as potential text queries and compute the 3D Intersection-over-Union (IoU) for individual class. We aggregate these measurements into dataset-level evaluation metrics, specifically mean IoU (mIoU) and mean accuracy (mAcc). Notably, unlike \cite{opengaussian}, which requires inputting all object categories at once and assigning each Gaussian primitive to the most relevant class, our method processes a single query at a time and generates segmentation results of the query. We argue that this approach better aligns with real-world applications.


\begin{table}[t]
    \vspace{0.5em}
  \scriptsize
  \caption{
    3D Open-Vocabulary Segmentation Benchmark. 
  }
  \label{object_under}
  \setlength{\arrayrulewidth}{1.2pt} 
  \centering
  \begin{tabularx}{0.47\textwidth}{cXXXXXX}
    \toprule 
      \multicolumn{1}{c}{} &
      \multicolumn{3}{c}{Replica} &
      \multicolumn{3}{c}{ScanNet} \\
    Methods & mAcc\textuparrow & mIoU\textuparrow & FPS\textuparrow 
    & mAcc\textuparrow & mIoU\textuparrow & FPS\textuparrow \\
    \midrule
     ConceptFusion*  & 28.02 & 11.49 & 0.49 & 21.22 & 10.64 & 0.52\\
    ConceptGraphs  & 38.21 & 18.16 & - & 44.28 & 23.94 & -\\
    LangSplat  & 10.32 & 4.17 & - & 8.18 & 2.93 & -\\
    OpenGaussian & 44.28 & 26.39 & - & 42.76 & 25.01 & -\\
    \textbf{Ours*} & \textbf{49.19} & \textbf{35.88} & \textbf{1.02} & \textbf{58.29} & \textbf{37.23} & \textbf{1.22}\\
  \bottomrule
    \multicolumn{6}{l}{* denotes methods running online}\\
  \end{tabularx}
    \vspace{-1.5em}
\end{table}

\noindent\textbf{Results.}
In Table \ref{object_under}, we present the quantitative metrics of all methods on two datasets. Our method achieves the best performance in both open-vocabulary 3D object segmentation accuracy and training efficiency. Specifically, our method outperforms the state-of-the-art 3DGS-based approach, OpenGaussian, with improvements of 9.5\% (Replica) and 12.2\% (ScanNet) in terms of mIoU performance, while operating in an online setting without the need for explicit scene segmentation. In comparison to point-cloud-based ConceptFusion, our method shows a significant improvement of over 20\% in mIoU in both datasets. We attribute this to the adaptive thresholding capability of our advanced AT-MLLM algorithm, whereas Conceptfusion employs a fixed threshold.

In Fig.\ref{exp1}, we present the qualitative metrics of our method across four scenes. As shown, since LangSplat only focuses on 2D rendering of objects from arbitrary viewpoints, its performance drops significantly when recovering semantic information into 3D space. While OpenGaussian enhances 3D understanding through 3D instance segmentation and semantic labeling but sacrifices granularity in open-vocabulary queries. For example, when querying for \textit{"Bins"} in a scene with two bins (third row). OpenGaussian fails to locate both instances as they are segmented into separate entities, and the model by default only retrieves the instance that best matches the query. Additionally, conveying quantitative object information in robotic interactions remains challenging.

\begin{figure*}[!t]
 \vspace{0.5em} 
\centering
	\includegraphics[width=17.5cm]{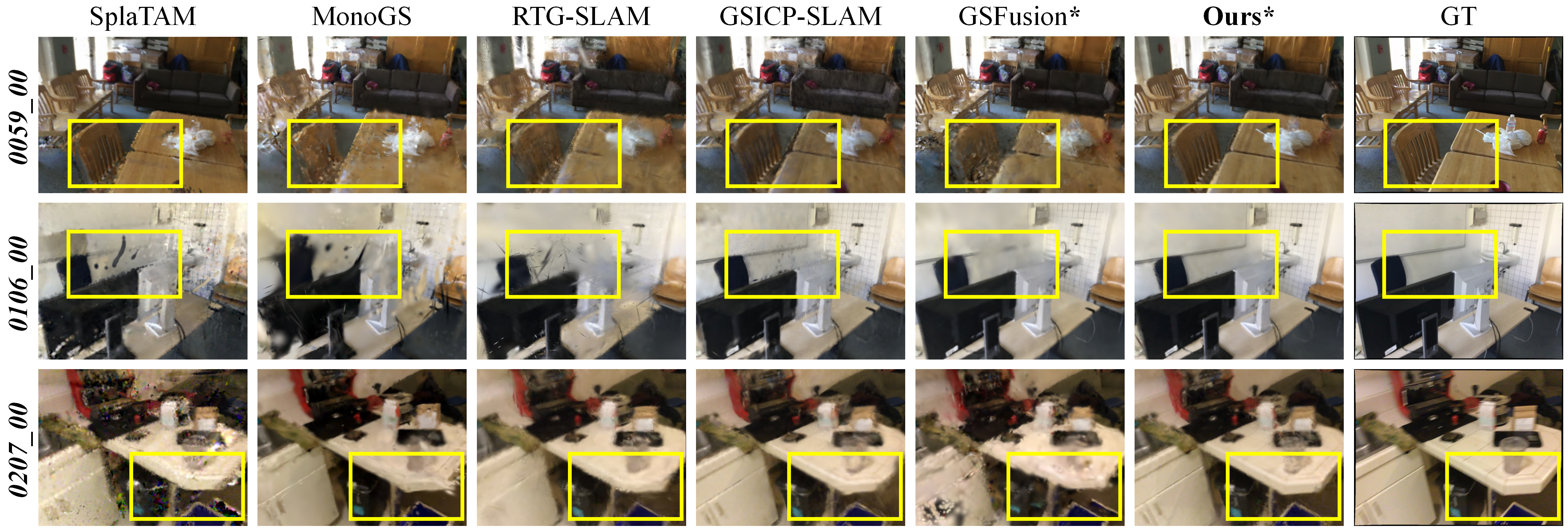}
	\caption{Qualitative rendering results from training views on the ScanNet dataset. Zoom in for a clearer view. * denotes that the method employs a hybrid scene representation. All methods in this comparison utilize ground-truth poses.}
	\label{exp2} 
    \vspace{-0.5em}
\end{figure*}

\begin{table}[!ht]
\centering
\caption{Quantitative Train View Rendering Performance on Scannet. (average performance on 6 scenes)}
\label{recons_scannet}
\begin{threeparttable}
\setlength{\arrayrulewidth}{1.2pt} 
\resizebox{0.48\textwidth}{!}{
\begin{tabular}{lcccccc}
    \toprule
    Method & PSNR \textuparrow & SSIM \textuparrow & LPIPS \textdownarrow & Depth L1 \textdownarrow & FPS \textuparrow \\
    \midrule
    MonoGS & 17.31 & 0.636 & 0.583 & 21.30 & 14.02 \\
    RTG-SLAM & 18.22 & 0.741 & 0.537 & 17.19 & 18.11 \\
    Splatam & 19.91 & 0.747 & 0.312 & 14.01 & 1.79 \\
    GS-Fusion & 21.51 & 0.774 & 0.251 & - & 24.68 \\
    GSICP-SLAM & 24.53 & 0.841 & 0.190 & 1.78 & \textbf{30.00} \\
    \textbf{Ours (w/o sem)} & \textbf{25.88} & \textbf{0.842} & \textbf{0.181} & \textbf{1.62} & 23.80 \\
    \bottomrule
\end{tabular}}
\end{threeparttable}
\vspace{-0.5em}
\end{table}

\subsection{Dense Scene Mapping} \label{Dense}
\noindent\textbf{Settings.} 1) Task: Given a posed RGB-D stream, our method can perform high-fidelity dense scene mapping. This task aims to evaluate the quality of geometric and appearance reconstruction of the scene, as well as the overall reconstruction efficiency. To ensure a fair comparison, we disable the semantics-related modules in our method.

2) Baselines: We select state-of-the-art dense mapping methods based on Gaussian splatting as comparative baselines, including SplaTAM\cite{splatam}, MonoGS\cite{monogs}, RTG-SLAM\cite{peng2024rtgslam} and GSICP-SLAM\cite{gsicp}. Additionally, we include GS-Fusion\cite{wei2024gsfusion}, which also utilizes a hybrid scene representation. For each method, we use their default parameters. 

3) Datasets: The experiments are carried out on 8 synthetic scenes from the Replica dataset and 6 real-world scenes from the ScanNet dataset, following the setup of \cite{splatam}.

4) Evaluation Metrics: We employ three standard metrics for rendering quality: PSNR, SSIM, and LPIPS. We also evaluated geometric reconstruction accuracy with depth L1 loss and system efficiency with mapping frame rate (FPS).

\vspace{0.5em}

\noindent\textbf{Results.} In Table \ref{recons_replica} and Table \ref{recons_scannet}, we present the quantitative reconstruction quality of our method in the Replica and ScanNet datasets. Our method achieves state-of-the-art performance in PSNR, SSIM, LPIPS, and Depth L1. In particular, the scene understanding module of our method is disabled in this evaluation, which explains the significant speed improvement compared to Table \ref{object_under}.

\begin{table}[!ht]
\centering
\caption{Quantitative Train View Rendering Performance on Replica. (average performance on 8 scenes)}
\label{recons_replica}
\begin{threeparttable}
\setlength{\arrayrulewidth}{1.2pt} 
\resizebox{0.48\textwidth}{!}{
\begin{tabular}{lcccccc}
    \toprule
    Method & PSNR \textuparrow & SSIM \textuparrow & LPIPS \textdownarrow & Depth L1 \textdownarrow & FPS \textuparrow \\
    \midrule
    MonoGS & 36.50 & 0.960 & 0.070 & 0.77 & 4.691 \\
    RTG-SLAM & 35.43 & 0.982 & 0.109 & 1.11 & 14.06 \\
    Splatam & 34.98 & 0.972 & 0.088 & 0.82 & 0.38 \\
    GS-Fusion & 34.78 & 0.948 & 0.055 & - & 18.41 \\
    GSICP-SLAM & 38.34 & 0.975 & 0.041 & 0.58 & \textbf{30.02} \\
    \textbf{Ours (w/o sem)} & \textbf{38.52} & \textbf{0.977} & \textbf{0.039} & \textbf{0.54} & 21.91 \\
    \bottomrule
\end{tabular}}
\end{threeparttable}
\vspace{-0.5em}
\end{table}

Compared to GS-Fusion, which initializes Gaussians with fixed parameters in each voxel grid, our method leverages a better initialization strategy that better captures the scene's geometric structure, leading to superior performance. Additionally, compared to our code-based approach GSICP-SLAM (the method upon which our work is developed), our performance improvements on ScanNet are more pronounced than on Replica. 
 We attribute this to the incorporation of our extra GS initialization and pruning mechanism, which leverages the TSDF to improve robustness when handling real-world scene data with depth noise. 

Furthermore, Fig. \ref{exp2} presents qualitative rendering results in four real-world scenes, highlighting the robustness of our method against motion blur and depth noise.

\begin{figure}[t]
\vspace{0.2em}
    \centering
    \includegraphics[width=0.96\linewidth]{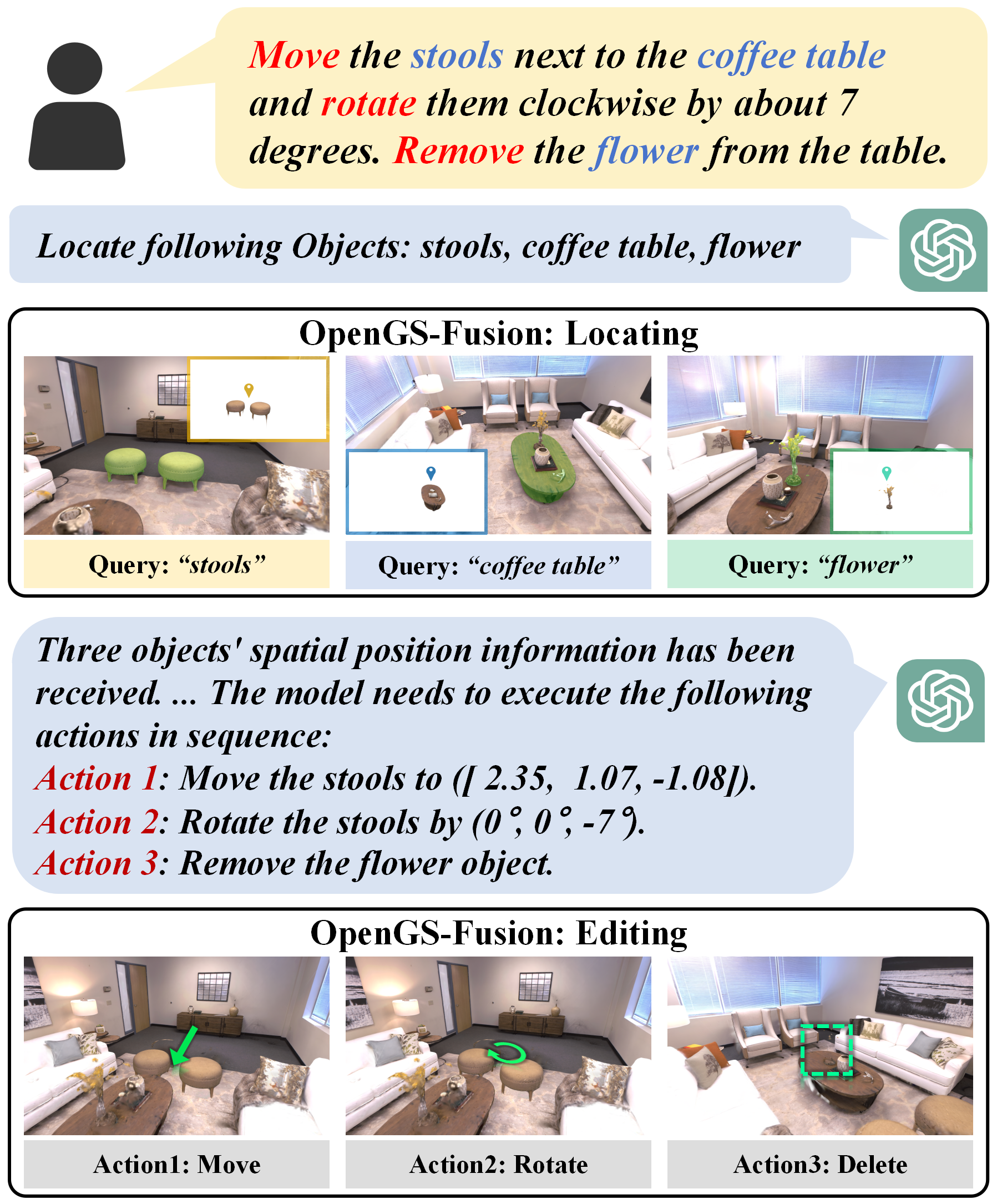}

    \caption{An Example of Language-Guided Scene Editing. OpenGS-Fusion accurately localizes objects in 3D space and efficiently executes language-guided scene modifications. The spatial position of objects is represented using a standardized format, encompassing nine dimensions, including center position coordinates, dimensions (length, width, height), and rotation angles.}
    \label{scene_edit}
    \vspace{-1.0em}
\end{figure}

\subsection{Language-guided scene editing} \label{exp_edit}
By integrating with large language models (LLMs) on the front end, OpenGS-Fusion empowers users to execute a diverse range of scene interaction tasks with enhanced efficiency and accuracy.  Consider a scenario where users provide natural language instructions to manipulate objects within a 3D environment. As illustrated in Fig. \ref{scene_edit}, the system first identifies and localizes the referenced objects, then sequentially executes spatial transformations such as repositioning, rotation, and object removal. This language-guided editing framework facilitates precise and immersive scene modifications, bridging the gap between human intentions and automated spatial reasoning.

\subsection{Ablation Experiments} \label{ablation}

To further validate our method, we conducted ablation studies focusing on the core parameters of our hybrid representation, specifically the voxel size of the TSDF and our AT-MLLM module. 

The results are shown in Table \ref{ablation_table}. \textit{1) Voxel Size: } In our method, the size of the voxel affects both the granularity of the semantic space and the initialization process of Gaussians. Reducing voxel size improves overall reconstruction quality and scene understanding capabilities. \textit{2) AT-MLLM Module: } This module adaptively adjusts thresholds using MLLM to filter out irrelevant primitives in the scene. Compared to a fixed threshold strategy (set at 0.6), our AT-MLLM module enhances the final scene understanding of mIoU by 17\% (at 5 cm of voxel size).

\begin{table}[!ht]
\caption{Ablation Study on ScanNet: Assessing Reconstruction Quality with Datasets from \ref{3D_open} and Scene Understanding with Datasets from \ref{Dense}}
\centering
\begin{threeparttable}
\resizebox{0.48\textwidth}{!}{
\begin{tabular}{cc|ccccc}
\hlineB{2.5}
\makecell{Voxel\\ size}           & \makecell{Adaptive \\ Threshold}   & PSNR$\uparrow$ & SSIM$\uparrow$ & \makecell{LPIPS}$\downarrow$ & \begin{tabular}[c]{@{}c@{}} mIoU(\%) \end{tabular} & \begin{tabular}[c]{@{}c@{}} mAcc(\%) \end{tabular} \\ \hlineB{2.5}
\multirow{2}{*}{10cm} & {\xmark} & \multirow{2}{*}{24.57} & \multirow{2}{*}{0.841} & \multirow{2}{*}{0.177}    & 14.26      &  23.50                                    \\
                     & {\cmark}    &  &  &    &     33.69      &  46.14                                                          \\ \hline
\multirow{2}{*}{5cm} & {\xmark} & \multirow{2}{*}{25.88} & \multirow{2}{*}{0.842} & \multirow{2}{*}{0.181}    & 19.77                                 & 34.81                                    \\
                     & {\cmark}    &  &  &    &     37.23      & 58.29                                                                   \\ \hlineB{2.5}
\end{tabular}
}
\label{ablation_table}
\end{threeparttable}
\end{table}

\subsection{Real-World Experiments} \label{real}
\begin{figure}[t]
\vspace{0.2em}
    \centering
    \includegraphics[width=0.98\linewidth]{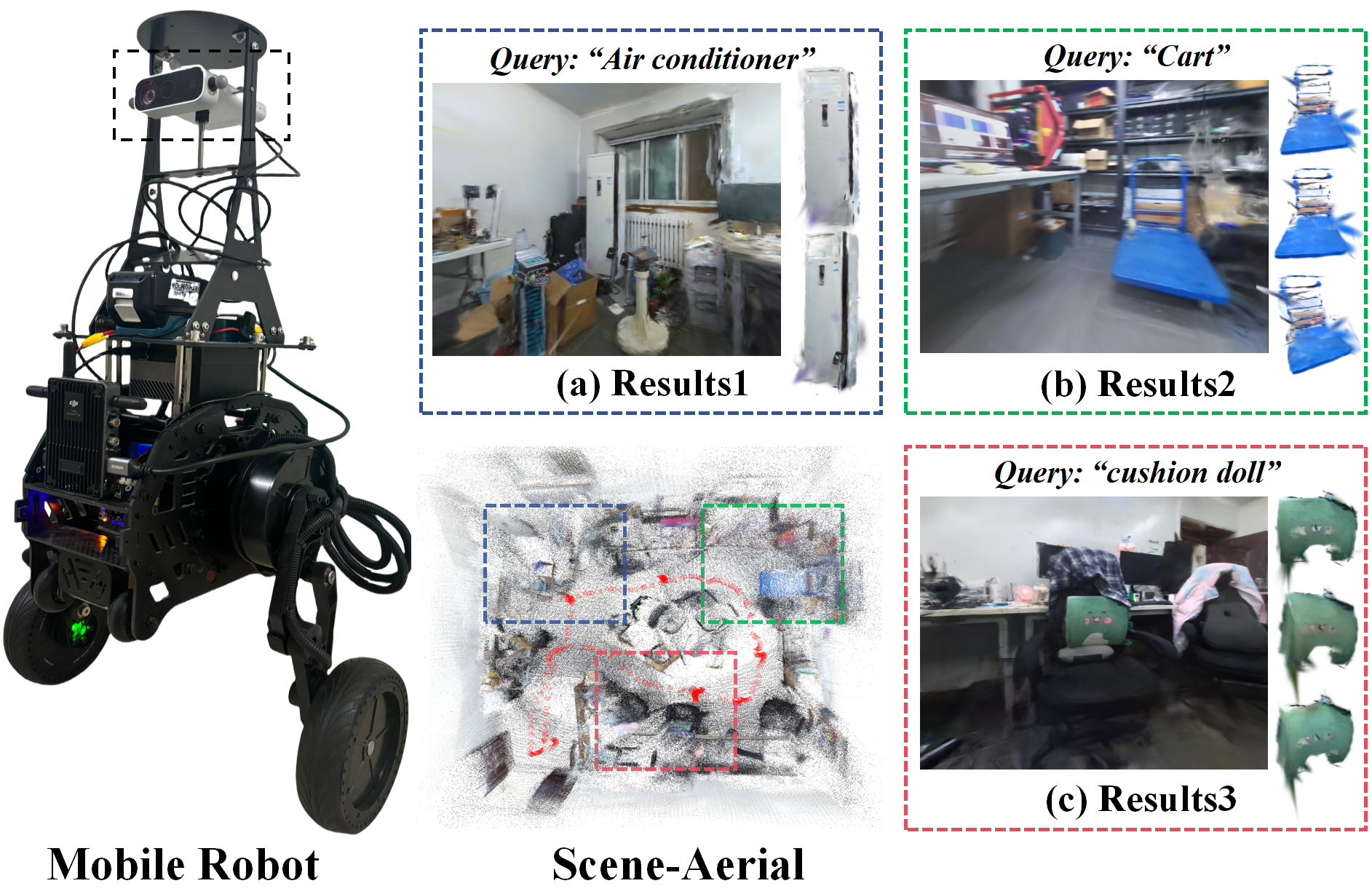}

    \caption{Reconstruction and scene understanding results on our self-captured dataset, demonstrated across three representative spatial locations.}
    \label{real-world}
    \vspace{-1.0em}
\end{figure}

In this section, we describe the practical implementation of OpenGS-Fusion for the reconstruction and understanding of indoor scenes using a mobile robotic device. Experiments were conducted on the Diablo platform equipped with a RGB-D camera system and an NVIDIA Jetson AGX Orin. The setup enables real-time image streaming and pose estimation via ORB-SLAM3. Our algorithm was deployed on a computing setup with an NVIDIA RTX 4090 GPU and an Intel Core i9-14900K CPU. To improve efficiency, MobileSAMv2\cite{zhang2023mobilesamv2} was used as the 2D segmentation model in our 2D Embedding Extractor model, enabling real-time image segmentation at more than 25 fps. Parallel processing on the server achieved an overall operational efficiency of 4 fps. Following \cite{monogs,wei2024gsfusion}, global optimization was performed after mapping to provide more realistic synthesized results. Figure \ref{real-world} shows the final results of scene reconstruction and object-level query within the scene.

\section{CONCLUSIONS}

We propose OpenGS-Fusion, an innovative dense mapping framework for open-vocabulary 3D scene understanding. It integrates 3D Gaussian representations with a Truncated Signed Distance Field  to fuse semantic information in an online manner and enhance modeling capabilities. By combining semantic features into the hybrid scene representation, our approach enables continuous scene modeling while maintaining high-quality geometric and semantic fidelity.  Additionally, we introduce MLLM-Assisted Adaptive Thresholding, a novel mechanism that refines 3D object localization by adaptively adjusting similarity thresholds based on MLLM's guidance. Through extensive evaluations, we demonstrate that OpenGS-Fusion outperforms existing methods in scene reconstruction quality and object-level understanding,  making it a strong candidate for real-world interaction tasks in various environments. However, our method currently relies on accurate pose estimation and faces limitations in query efficiency. Future work will explore how to leverage hybrid scene representation for pose estimation and investigate lightweight MLLMs specifically designed for image retrieval tasks to further enhance performance.

\section*{APPENDIX}
\label{append}
In this section, we delve into the implementation details of our method. For 2D feature extraction, we utilize SAM\cite{sam} and the default CLIP\cite{openclip} model in our experiments, following the approach adopted by other methods. For scene representation, we set the leaf voxel size of the VDB structure to 5cm and the truncated value to 7cm. We adhere to the GS scene representation settings outlined in \cite{gsicp}.  For our AT-MLLM method, we set the initial threshold $\delta_1 = 0.8$, select $u=3$ keyframes, and use a threshold window $\xi=0.2$ per iteration. We set the MLLM as gpt-4o-mini, which requires an average of 5 seconds per iteration. For mapping, we select a keyframe every 10 frames on Replica and every 8 frames on ScanNet. During optimization, frames are randomly selected from the keyframes window.

\bibliographystyle{Bibliography/IEEEtran}
\bibliography{Bibliography/IROS}

\begin{thebibliography}{10}
\providecommand{\url}[1]{#1}
\csname url@samestyle\endcsname
\providecommand{\newblock}{\relax}
\providecommand{\bibinfo}[2]{#2}
\providecommand{\BIBentrySTDinterwordspacing}{\spaceskip=0pt\relax}
\providecommand{\BIBentryALTinterwordstretchfactor}{4}
\providecommand{\BIBentryALTinterwordspacing}{\spaceskip=\fontdimen2\font plus
\BIBentryALTinterwordstretchfactor\fontdimen3\font minus \fontdimen4\font\relax}
\providecommand{\BIBforeignlanguage}[2]{{%
\expandafter\ifx\csname l@#1\endcsname\relax
\typeout{** WARNING: IEEEtran.bst: No hyphenation pattern has been}%
\typeout{** loaded for the language `#1'. Using the pattern for}%
\typeout{** the default language instead.}%
\else
\language=\csname l@#1\endcsname
\fi
#2}}
\providecommand{\BIBdecl}{\relax}
\BIBdecl

\bibitem{conceptfusion}
K.~Jatavallabhula\emph{,~et~al.}, ``Conceptfusion: Open-set multimodal 3d mapping,'' in \emph{Proc. of Robotics: Science and Systems (RSS)}, 2023.

\bibitem{lerf2023}
J.~Kerr\emph{,~et~al.}, ``Lerf: Language embedded radiance fields,'' in \emph{Proc. of International Conference on Computer Vision (ICCV)}, 2023.

\bibitem{open-fusion}
K.~Yamazaki\emph{,~et~al.}, ``Open-fusion: Real-time open-vocabulary 3d mapping and queryable scene representation,'' in \emph{Proc. of IEEE International Conference on Robotics and Automation (ICRA)}, pp. 9411--9417, 2024.

\bibitem{opengaussian}
Y.~Wu\emph{,~et~al.}, ``Opengaussian: Towards point-level 3d gaussian-based open vocabulary understanding,'' in \emph{Proc. of Advances in Neural Information Processing Systems}, 2024.

\bibitem{openclip}
A.~Radford\emph{,~et~al.}, ``Learning transferable visual models from natural language supervision,'' in \emph{Proc. of International Conference on Machine Learning (ICML)}, 2021.

\bibitem{gaussian_grouping}
M.~Ye\emph{,~et~al.}, ``Gaussian grouping: Segment and edit anything in 3d scenes,'' in \emph{Proc. of European Conference on Computer Vision (ECCV)}, 2024.

\bibitem{VR-GS}
Y.~Jiang\emph{,~et~al.}, ``Vr-gs: A physical dynamics-aware interactive gaussian splatting system in virtual reality,'' in \emph{Proc. of ACM SIGGRAPH 2024 Conference Papers}, 2024.

\bibitem{VLM}
C.~Huang\emph{,~et~al.}, ``Visual language maps for robot navigation,'' in \emph{Proc. of IEEE International Conference on Robotics and Automation (ICRA)}, pp. 10\,608--10\,615, 2023.

\bibitem{clipfields}
N.~M.~M. Shafiullah\emph{,~et~al.}, ``Clip-fields: Weakly supervised semantic fields for robotic memory,'' \emph{arXiv preprint arXiv: Arxiv-2210.05663}, 2022.

\bibitem{lerftogo}
A.~Rashid\emph{,~et~al.}, ``Language embedded radiance fields for zero-shot task-oriented grasping,'' in \emph{Conference on Robot Learning (CoRL)}, 2023.

\bibitem{gaussiangrasper}
Y.~Zheng\emph{,~et~al.}, ``Gaussiangrasper: 3d language gaussian splatting for open-vocabulary robotic grasping,'' \emph{arXiv preprint arXiv:2403.09637}, 2024.

\bibitem{langsplat}
M.~Qin\emph{,~et~al.}, ``Langsplat: 3d language gaussian splatting,'' in \emph{Proc. of IEEE/CVF Conference on Computer Vision and Pattern Recognition (CVPR)}, pp. 20\,051--20\,060, 2024.

\bibitem{legaussian}
J.-C. Shi\emph{,~et~al.}, ``Language embedded 3d gaussians for open-vocabulary scene understanding,'' in \emph{Proc. of IEEE/CVF Conference on Computer Vision and Pattern Recognition (CVPR)}, pp. 5333--5343, 2024.

\bibitem{feature-3dgs}
S.~Zzhou\emph{,~et~al.}, ``Feature 3dgs: Supercharging 3d gaussian splatting to enable distilled feature fields,'' in \emph{Proc. of IEEE/CVF Conference on Computer Vision and Pattern Recognition}, pp. 21\,676--21\,685, 2024.

\bibitem{FMGS}
X.~Zuo\emph{,~et~al.}, ``Fmgs: Foundation model embedded 3d gaussian splatting for holistic 3d scene understanding,'' \emph{Int. J. Comput. Vision}, vol. 133, no.~2, pp. 611--627, 2024.

\bibitem{uni-fusion}
Y.~Yuan\emph{,~et~al.}, ``Uni-fusion: Universal continuous mapping,'' \emph{IEEE Transactions on Robotics}, vol.~40, pp. 1373--1392, 2024.

\bibitem{splatam}
N.~Keetha\emph{,~et~al.}, ``Splatam: Splat, track and map 3d gaussians for dense rgb-d slam,'' in \emph{Proc. of the IEEE/CVF Conference on Computer Vision and Pattern Recognition}, 2024.

\bibitem{3d-ovs}
K.~Liu\emph{,~et~al.}, ``Weakly supervised 3d open-vocabulary segmentation,'' in \emph{Proc. of Advances in Neural Information Processing Systems}, 2023.

\bibitem{nef2}
Y.~Bhalgat\emph{,~et~al.}, ``N2f2: Hierarchical scene understanding with nested neural feature fields,'' in \emph{Proc. of European Conference on Computer Vision (ECCV)}, pp. 197--214, 2024.

\bibitem{LEGS}
J.~Yu\emph{,~et~al.}, ``Language-embedded gaussian splats (legs): Incrementally building room-scale representations with a mobile robot,'' in \emph{2024 IEEE/RSJ International Conference on Intelligent Robots and Systems (IROS)}, pp. 13\,326--13\,332, 2024.

\bibitem{3dgs}
B.~Kerbl\emph{,~et~al.}, ``3d gaussian splatting for real-time radiance field rendering,'' \emph{ACM Transactions on Graphics}, vol.~42, no.~4, pp. 1--14, 2023.

\bibitem{tiger}
T.~Xu\emph{,~et~al.}, ``Tiger: Text-instructed 3d gaussian retrieval and coherent editing,'' \emph{arXiv preprint arXiv:2405.14455}, 2024.

\bibitem{clipgs}
G.~Liao\emph{,~et~al.}, ``Clip-gs: Clip-informed gaussian splatting for real-time and view-consistent 3d semantic understanding,'' \emph{arXiv preprint arXiv:2404.14249}, 2024.

\bibitem{gsicp}
S.~Ha\emph{,~et~al.}, ``Rgbd gs-icp slam,'' in \emph{Proc. of European Conference on Computer Vision (ECCV)}, pp. 180--197, 2024.

\bibitem{vdbfusion}
I.~Vizzo\emph{,~et~al.}, ``Vdbfusion: Flexible and efficient tsdf integration of range sensor data,'' \emph{Sensors}, vol.~22, no.~3, p. 1296, 2022.

\bibitem{gpt}
J.~Achiam\emph{,~et~al.}, ``Gpt-4 technical report,'' \emph{arXiv preprint arXiv:2303.08774}, 2023.

\bibitem{nerf}
B.~Mildenhall\emph{,~et~al.}, ``Nerf: Representing scenes as neural radiance fields for view synthesis,'' \emph{Communications of the ACM}, vol.~65, no.~1, pp. 99--106, 2021.

\bibitem{loopsplta}
L.~Zhu\emph{,~et~al.}, ``Loopsplat: Loop closure by registering 3d gaussian splats,'' in \emph{International Conference on 3D Vision (3DV)}, 2025.

\bibitem{monogs}
H.~Matsuki\emph{,~et~al.}, ``{G}aussian {S}platting {SLAM},'' in \emph{Proc. of the IEEE/CVF Conference on Computer Vision and Pattern Recognition}, 2024.

\bibitem{cgslam}
J.~Hu\emph{,~et~al.}, ``Cg-slam: Efficient dense rgb-d slam in a consistent uncertainty-aware 3d gaussian field,'' in \emph{Proc. of the European Conference on Computer Vision (ECCV)}, pp. 93--112, 2024.

\bibitem{wei2024gsfusion}
J.~Wei\emph{,~et~al.}, ``Gsfusion: Online rgb-d mapping where gaussian splatting meets tsdf fusion,'' \emph{IEEE Robotics and Automation Letters}, 2024.

\bibitem{sgsslam}
M.~Li\emph{,~et~al.}, ``Sgs-slam: Semantic gaussian splatting for neural dense slam,'' in \emph{Proc. of the European Conference on Computer Vision (ECCV)}, pp. 163--179, 2024.

\bibitem{semgauss}
S.~Zhu\emph{,~et~al.}, ``Semgauss-slam: Dense semantic gaussian splatting slam,'' \emph{arXiv preprint arXiv:2403.07494}, 2024.

\bibitem{regionclip}
Y.~Zhong\emph{,~et~al.}, ``Regionclip: Region-based language-image pretraining,'' in \emph{Proc. of the IEEE/CVF Conference on Computer Vision and Pattern Recognition (CVPR)}, pp. 16\,793--16\,803, 2022.

\bibitem{zhou2022maskclip}
C.~Zhou\emph{,~et~al.}, ``Extract free dense labels from clip,'' in \emph{Proc. of the European Conference on Computer Vision (ECCV)}, 2022.

\bibitem{Peng2023OpenScene}
S.~Peng\emph{,~et~al.}, ``Openscene: 3d scene understanding with open vocabularies,'' in \emph{Proc. of the IEEE/CVF Conference on Computer Vision and Pattern Recognition (CVPR)}, 2023.

\bibitem{nguyen2023open3dis}
P.~D.~A. Nguyen\emph{,~et~al.}, ``Open3dis: Open-vocabulary 3d instance segmentation with 2d mask guidance,'' in \emph{Proc. of the IEEE/CVF Conference on Computer Vision and Pattern Recognition (CVPR)}, 2024.

\bibitem{OV-NeRF}
G.~Liao\emph{,~et~al.}, ``Ov-nerf: Open-vocabulary neural radiance fields with vision and language foundation models for 3d semantic understanding,'' \emph{IEEE Transactions on Circuits and Systems for Video Technology}, vol.~34, no.~12, pp. 12\,923--12\,936, 2024.

\bibitem{ovmapping}
M.~Tie\emph{,~et~al.}, ``Ov-mapping: Online open-vocabulary mapping with neural implicit representation,'' in \emph{Proc. of the European Conference on Computer Vision (ECCV)}, A.~Leonardis\emph{,~et~al.}, Eds., 2025.

\bibitem{fastlgs}
Y.~Ji\emph{,~et~al.}, ``Fastlgs: Speeding up language embedded gaussians with feature grid mapping,'' in \emph{Proc. of AAAI Conference on Artificial Intelligence}, 2025.

\bibitem{gu2024egolifter}
Q.~Gu\emph{,~et~al.}, ``Egolifter: Open-world 3d segmentation for egocentric perception,'' in \emph{Proc. of the European Conference on Computer Vision (ECCV)}, pp. 382--400, Milan, Italy, 2024.

\bibitem{zhang2023mobilesamv2}
C.~Zhang\emph{,~et~al.}, ``Mobilesamv2: Faster segment anything to everything,'' \emph{arXiv preprint arXiv:2312.09579}, 2023.

\bibitem{sam}
A.~Kirillov\emph{,~et~al.}, ``Segment anything,'' \emph{arXiv:2304.02643}, 2023.

\bibitem{dbscn}
M.~Ester\emph{,~et~al.}, ``A density-based algorithm for discovering clusters in large spatial databases with noise,'' in \emph{Proc. of the Second International Conference on Knowledge Discovery and Data Mining}, pp. 226--231.\hskip 1em plus 0.5em minus 0.4em\relax AAAI Press, 1996.

\bibitem{conceptgraph}
Q.~Gu\emph{,~et~al.}, ``Conceptgraphs: Open-vocabulary 3d scene graphs for perception and planning,'' in \emph{Proc. of IEEE International Conference on Robotics and Automation (ICRA)}, pp. 5021--5028, 2024.

\bibitem{replica19arxiv}
J.~Straub\emph{,~et~al.}, ``The {R}eplica dataset: A digital replica of indoor spaces,'' \emph{arXiv preprint arXiv:1906.05797}, 2019.

\bibitem{dai2017scannet}
A.~Dai\emph{,~et~al.}, ``Scannet: Richly-annotated 3d reconstructions of indoor scenes,'' in \emph{Proc. Computer Vision and Pattern Recognition (CVPR), IEEE}, 2017.

\bibitem{peng2024rtgslam}
Z.~Peng\emph{,~et~al.}, ``Rtg-slam: Real-time 3d reconstruction at scale using gaussian splatting,'' 2024.

\end{thebibliography}

\end{document}